\documentclass[letterpaper, 10 pt, conference]{ieeeconf}  % Comment this line out if you need a4paper

\IEEEoverridecommandlockouts                              % This command is only needed if 
\usepackage{cite}
\usepackage{url}
\usepackage{amsmath,amssymb,amsfonts}
\usepackage{tikz}
\usepackage{graphicx}
\usepackage{textcomp}
\usepackage{xcolor}
\usepackage[ruled,vlined]{algorithm2e}
\SetArgSty{textnormal}
\usepackage{float}

\usepackage{algpseudocode}
\usepackage{comment}
\usepackage{todonotes}  % use option disable for final version

\usepackage[]{xcolor}

\usepackage{booktabs}
\usepackage{multirow}
\usepackage[usestackEOL]{stackengine}
\def\BibTeX{{\rm B\kern-.05em{\sc i\kern-.025em b}\kern-.08em
    T\kern-.1667em\lower.7ex\hbox{E}\kern-.125emX}}
\usepackage[firstpage=true,color=black,placement=top,scale=1,opacity=1,vshift=-0.3cm]{background}
\SetBgContents{
 \begin{minipage}{2.3\linewidth}
 \small
\textcopyright 2023 IEEE.  Personal use of this material is permitted.  Permission from IEEE must be obtained for all other uses, in any current or future media, including reprinting/republishing this material for advertising or promotional purposes, creating new collective works, for resale or redistribution to servers or lists, or reuse of any copyrighted component of this work in other works. A definitive version was published in \textbf{23rd IEEE International Conference on Nanotechnology 2023 (IEEE-NANO)} and is available at \url{https://doi.org/10.1109/nano58406.2023.10231169}
 \end{minipage}
 }
\title{\LARGE \bf
Instance Segmentation of Dislocations in TEM Images
}

\author{Karina Ruzaeva$^{1}$ \qquad Kishan Govind$^{1}$ \qquad Marc Legros$^{2}$ \qquad Stefan Sandfeld$^{1,3,\star}$% <-this % stops a space
\thanks{This work was supported by the European Union's horizon 2020 research and innovation program under grant agreement no 823717 (ESTEEM3) and by the European Research Council through the ERC Grant Agreement No. 759419 MuDiLingo (“A Multiscale Dislocation Language for Data-Driven Materials Science”).}% <-this % stops a space
\thanks{$^{1}$Materials Data Science and Informatics (IAS-9), Forschungszentrum Jülich GmbH, Jülich, 52425, Germany}%
\thanks{$^{2}$ CEMES-CNRS, Toulouse, 31055, France}%
\thanks{$^{3}$ Faculty of Georesources and Materials Engineering, RWTH Aachen University, Aachen, 52068, Germany}%
\thanks{$^{\star}$Corresponding author:
{\tt\small s.sandfeld@fz-juelich.de}}
}

\begin{document}

\maketitle
\thispagestyle{empty}
\pagestyle{empty}
\begin{abstract}
    Quantitative Transmission Electron Microscopy (TEM) during in-situ straining experiment is 
    able to reveal the motion of dislocations -- linear defects in the crystal lattice of metals. 
    In the domain of materials science, the knowledge about the location and movement of 
    dislocations is important for creating novel materials with superior properties. A long-standing
    problem, however, is to identify the position and extract the shape of dislocations, which 
    would ultimately help to create a digital twin of such materials.
    
    In this work, we quantitatively compare  state-of-the-art instance segmentation methods, 
    including Mask R-CNN and YOLOv8.
    The dislocation masks as the results of the instance segmentation are converted to mathematical
    lines, enabling quantitative analysis of dislocation length and geometry -- important
    information for the domain scientist, which we then propose to include as a novel length-aware 
    quality metric for estimating the network performance. Our segmentation pipeline shows 
    a high accuracy suitable for all domain-specific, further post-processing. Additionally,
    our physics-based metric turns out to perform much more consistently than typically used
    pixel-wise metrics.
\end{abstract}

\section{Introduction}
Dislocations are crystalline defects that constitute themselves on the nano-scale of 
metallic materials and alloys. Despite their small size, dislocations are able to
strongly influence many material properties on much larger scales. Recent 
technological advancements have greatly enhanced the capabilities of transmission electron microscopy (TEM), which allows to generate a large amount of image data. Therefore, the microscopy community is in urgent need of automatic high-throughput data-processing pipelines for materials characterization.

As many of the typical images are difficult to segment (e.g., image data from two different experiments look very different, the lightning conditions during every experiment are constantly changing, and dislocations strongly change their intensity during motion), the current state-of-the-art for identifying 
defects in TEM images still mainly relies on manual analysis, which is a laborious and time-consuming process. This strongly limits the analysis throughput and makes it a bottleneck in the experimental analysis.
Additionally, automated image analysis can provide more consistent and objective results compared to manual analysis, which can be prone to human errors and subjectivity. Therefore, automatic image analysis has become increasingly popular in various research fields and is an essential tool for analyzing large volumes of image data efficiently and accurately. The current work is a step in the direction of automated,
high-throughput data analysis with a particular focus on high accuracy such that the resulting
data can be easily used for further processing or even as input for nano-scale simulation methods.

\subsection{Machine Learning-based segmentation}
Image segmentation methods can be split into two major categories: semantic and instance segmentation.
Semantic segmentation (e.g., \cite{guo2018review,yu2018methods}) treats multiple objects within a single category as one entity. It performs a binary segmentation in case of having two classes (foreground and background), making it an excellent choice for the segmentation task where the objects are sparsely distributed. However, it requires post-processing to split any merged or overlapping objects.
On the other hand, instance segmentation \cite{hafiz2020survey} assigns a separate label to each individual instance of an object in an image within one class. In other words, it not only performs the segmentation task but also distinguishes between different instances of the same class.

%\subsection{ML in Material science (Prior work)}
Machine learning (ML) is a rapidly developing field that has shown great potential in various scientific disciplines, including electron microscopy, and it is a powerful tool for identifying and analyzing defects in materials at the nano-scale. 

Examples of successful ML application includes instance segmentation and classification of screw or edge dislocations~\cite{Nguyen2023} and irradiation-induced defects~\cite{Jacobs2022} or of grain structures~\cite{Trampert2021}; semantic segmentation has been performed on defects in steel~\cite{Roberts2019}, as well as ML-based semantic segmentation of the atomically resolved TEM images~\cite{Sadre2021}.
Dislocations in TEM images appear as thin lines that can resemble cracks or fracture networks. Therefore, it might be possible to adapt the techniques used for segmenting cracks in concrete surfaces, i.e., \cite{Hsu2021,Kim2020} to segment dislocations in TEM images.
\subsection{Data acquisition}
In-situ TEM straining experiments were performed on a single-phase face-centered, cubic CoCrFeMnNi alloy (also known as “Cantor alloy”). Samples were prepared as electron-transparent, stretchable thin foils following the method described in \cite{Oliveros2021,Zhang2022}. Each straining experiment was performed at 96\,K using a Gatan 671 straining holder in a JEOL 2010, which was operated at 200\,kV. The videos were captured at 22 images/s using a Megaview III SIS CCD camera and stored in MPEG-4 format on a hard drive. In this work, we selected more than ten typical video sequences, ranging from 10 to 200\,s that show discernible dislocation pile-ups gliding in (111) planes, which is expected from fcc-structured metals and alloys. Some sequences are not published and currently being analyzed using, to a certain extent, the presented method, while others were already manually analyzed and published. For more experimental details, we refer to \cite{Oliveros2021}. The ultimate goal of such an approach is to acquire sufficient knowledge about individual and collective dislocation dynamics to understand the macroscopic mechanical behavior of the alloy they move in. In the case of High Entropy Alloys (in which these sequences were captured), a long-standing goal is to establish the difference between the dislocation behavior in such chemically disordered structures and more classical alloys such as austenitic steels. 
\begin{figure}[htbp]
    \centering
    \begin{tikzpicture}
    \node[anchor=south west,inner sep=0] (Bild) at (0,0)
    {\includegraphics[width=.85\columnwidth]{ 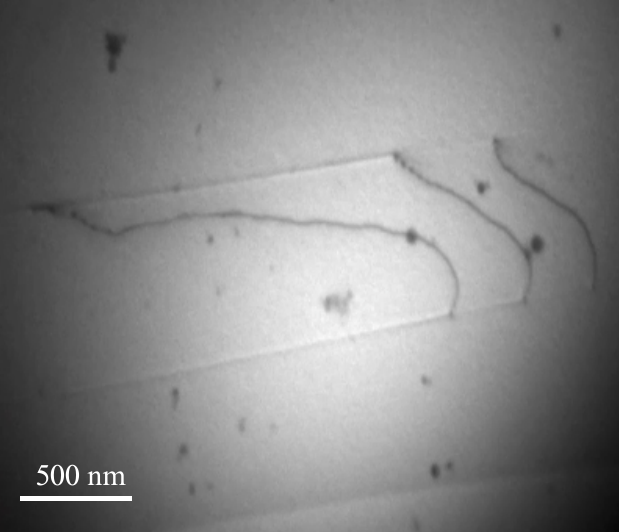}};
    \begin{scope}[x=(Bild.south east),y=(Bild.north west)]
    \node[color=black] at (0.6,0.53) {\textbf{1}};
    \node[color=black] at (0.7,0.6) {\textbf{2}};
    \node[color=black] at (0.8,0.66) {\textbf{3}};
    \node[color=black] at (0.55,0.3) {\textbf{II}};
    \node[color=black] at (0.4,0.72) {\textbf{I}};

    \end{scope}
    \end{tikzpicture}
    \caption{TEM image of the in-situ straining experiment, that illustrates the dislocation pile-up, that consists of three dislocations (1,2,3) to be identified and segmented, and the slip traces (I,II) to be ignored.}
    \label{3dis}
\end{figure}
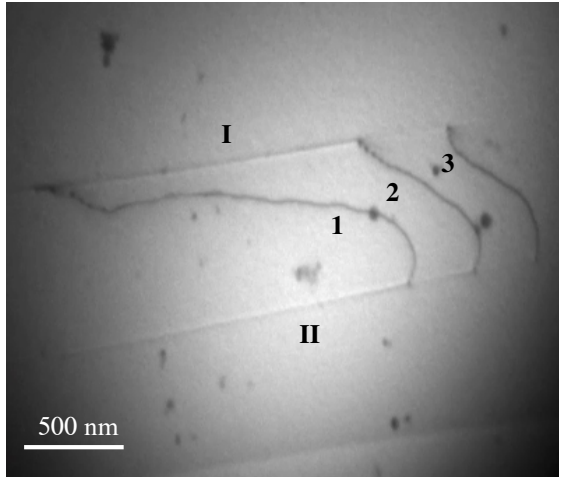

%The choice of the experiments and the corresponding videos is such as to demonstrate the significant
%variance of typical image data.
\subsection{Annotation process}  
The acquired videos of the straining experiments were randomly sampled, and 230 video ``snapshots'' (or ``frames'', e.g., shown in Figure~\ref{3dis}) were manually annotated for the instance segmentation in the annotation software \cite{labelme} by a domain expert. The annotated dataset was split into training (70\%) and validation datasets (30\%) and converted to the COCO and YOLO annotation formats using the software in~\cite{roboflow}.

\subsection{Challenges}
Dislocations are observed in the TEM as  narrow, curved and elongated objects. Instance segmentation of these brings certain challenges:

\begin{itemize}
    \item Our experimental data is not matching or even similar to any of the existing benchmark datasets, which typically represent "real-world" data, making transfer learning~\cite{Hosna2022} impossible. Creating annotations for instance segmentation can be  time-consuming and often requires annotation by the domain experts.
    \item In segmentation frameworks such as YOLOv8 and Mask R-CNN, a binary cross-entropy loss function is used as the mask loss. This loss is used to compare the predicted masks generated by the network to the ground truth masks provided in the training data. While the loss may work fairly well for, e.g., axis-aligned and round-shaped objects, it may have difficulties in mask prediction of dislocations that may exhibit a range of different orientations and which in many of the investigated frames were approximately "diagonally" oriented. The reason for this difficulty is the fact that the mask of diagonally oriented objects becomes sparsely represented in the bounding box, leading to class (object vs. background) imbalance. Therefore, in this case, the network may be biased towards the majority class (background) and neglect the minority class (dislocation). From a physical point of view, a suitable segmentation framework should be rotational-invariant concerning the orientation of the dislocations.
    \item Non-Maximum Suppression (NMS) is a technique used in object detection and instance segmentation algorithms to eliminate duplicate detections of the same object based on the degree of overlap between the proposals. The majority of state-of-the-art instance segmentation frameworks apply NMS at the bounding box level. Again, in the case of  diagonally oriented dislocations, a great overlap of the bounding boxes can be observed (cf. Figure~\ref{NMS}). Therefore, two closely located dislocations often get merged (or one with the lower confidence gets omitted) with a low NMS parameter (as illustrated in Figure~\ref{NMS}(B)). Increasing the NMS parameter has the opposite effect and results in duplicate detection of the same objects, as can be observed in Figure~\ref{NMS}. 
\end{itemize}
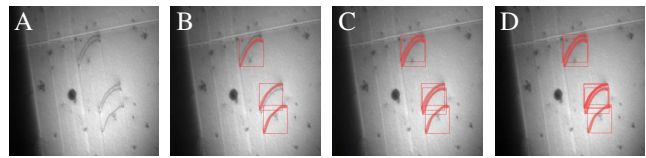
\begin{figure}[ht]
    \centering
    \begin{tikzpicture}
    \node[anchor=south west,inner sep=0] (Bild) at (0,0)
    {\includegraphics[width=.23\columnwidth]{ 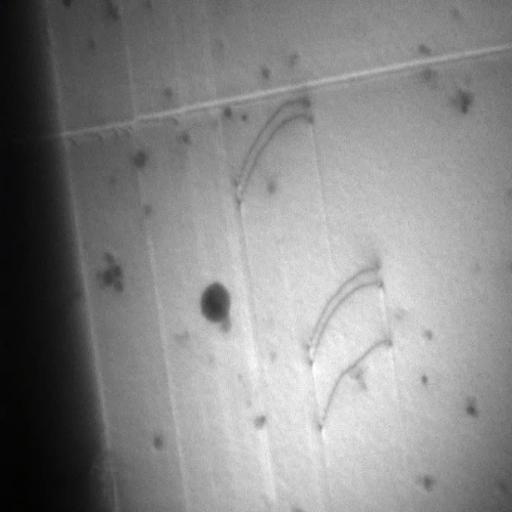}};
    \begin{scope}[x=(Bild.south east),y=(Bild.north west)]
    \node[color=white] at (0.1,0.9) {A};
    \end{scope}
    \end{tikzpicture}
    \begin{tikzpicture}
    \node[anchor=south west,inner sep=0] (Bild) at (0,0)
    {\includegraphics[width=.23\columnwidth]{ 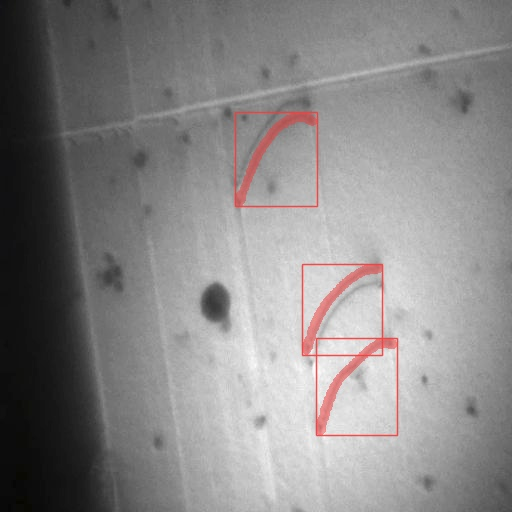}};
    \begin{scope}[x=(Bild.south east),y=(Bild.north west)]
    \node[color=white] at (0.1,0.9) {B};
    \end{scope}
    \end{tikzpicture}
    \begin{tikzpicture}
    \node[anchor=south west,inner sep=0] (Bild) at (0,0)
    {\includegraphics[width=.23\columnwidth]{ 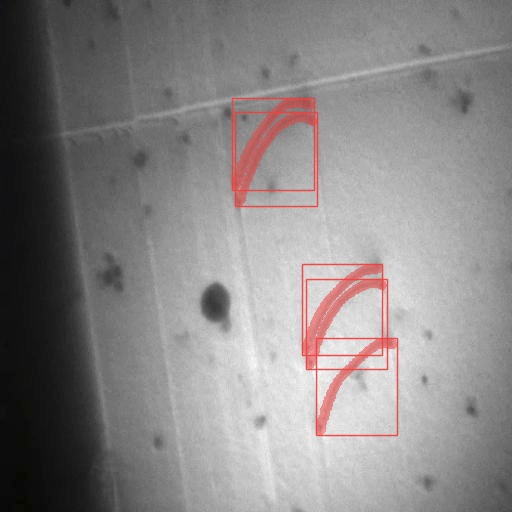}};
    \begin{scope}[x=(Bild.south east),y=(Bild.north west)]
    \node[color=white] at (0.1,0.9) {C};
    \end{scope}
    \end{tikzpicture}
    \begin{tikzpicture}
    \node[anchor=south west,inner sep=0] (Bild) at (0,0)
    {\includegraphics[width=.23\columnwidth]{ 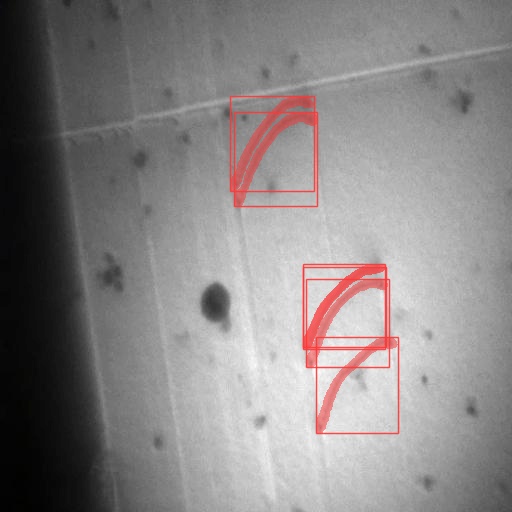}};
    \begin{scope}[x=(Bild.south east),y=(Bild.north west)]
    \node[color=white] at (0.1,0.9) {D};
    \end{scope}
    \end{tikzpicture}
    \caption{The original image (A) and the results of the instance segmentation (B, C and D) with different NMS parameters (0.6, 0.8 and 0.95, respectively)}
    \label{NMS}
\end{figure}

\section{Methods}

\subsection{ML-based instance segmentation}

In this work, we compare the performance of two commonly used instance segmentation frameworks, Mask R-CNN and YOLOv8.

\subsubsection{Mask R-CNN}
Mask R-CNN \cite{He2017} is a popular deep learning model for instance segmentation. It was created by adding a mask branch to the Faster R-CNN \cite{Ren2015} architecture, which allows the model to output a binary mask for each detected object in addition to the bounding box and class label.
Mask R-CNN uses anchors as part of the Region Proposal Network (RPN) to generate candidate object bounding boxes. The anchors are predefined boxes of different sizes and aspect ratios that are centered at every pixel location in a feature map. RPN predicts the offsets and scale factors of the anchors to generate the final set of object proposals\cite{Ren2015}. In this paper, we use the MMDetection implementation of the network \cite{mmdetection}.

\subsubsection{YOLOv8}
YOLOv8 is a state-of-the-art object detection and image segmentation model created by Ultralytics~\cite{Jocher2023}. YOLOv8 is an anchor-free model. This means it directly 
predicts the center of an object instead of the offset from a known anchor box. Anchor-free detection reduces the number of box predictions, which speeds up Non-Maximum Suppression \cite{roboflow}. 
To improve the robustness and generalization of the approach, Ultralytics provides basic augmentation techniques, e.g, image scaling and flipping, as well as mosaic augmentation, first introduced in \cite{Bochkovskiy2020}. 

\subsection{Calculation of dislocation length from the segmentation masks}
To produce training data for the deep learning models used in our work, experts manually annotated dislocations in TEM images by drawing their contours. However, there was no consistency in how wide the masks obtained from these contours were drawn, 
leading to variability in the width of predicted masks by the network. As materials scientists are primarily interested in properties such as length, orientation, and geometry of the dislocations (as e.g., used for data mining in \cite{STEINBERGER2023111830}), this variability in mask width is undesired.
To address this problem, we performed Lee skeletonization~\cite{au2008skeleton} of the predicted masks to reduce the masks to one-pixel width representation, making it easier to calculate the length and geometry of the dislocations. In particular, the dislocation length can be found either as a rough estimate in proportion to the sum of the pixels of the obtained skeleton or, more accurately, from fitting a polygon or spline to the skeleton. 

\begin{figure}[htbp]
	\centering
    \hspace{-0.25cm}
    \stackinset{c}{}{t}{-.15in}{\small Original Image}{
    \begin{tikzpicture}
    \node[anchor=south west,inner sep=0] (Bild) at (0,0)
    {\includegraphics[width=.3\columnwidth]{ 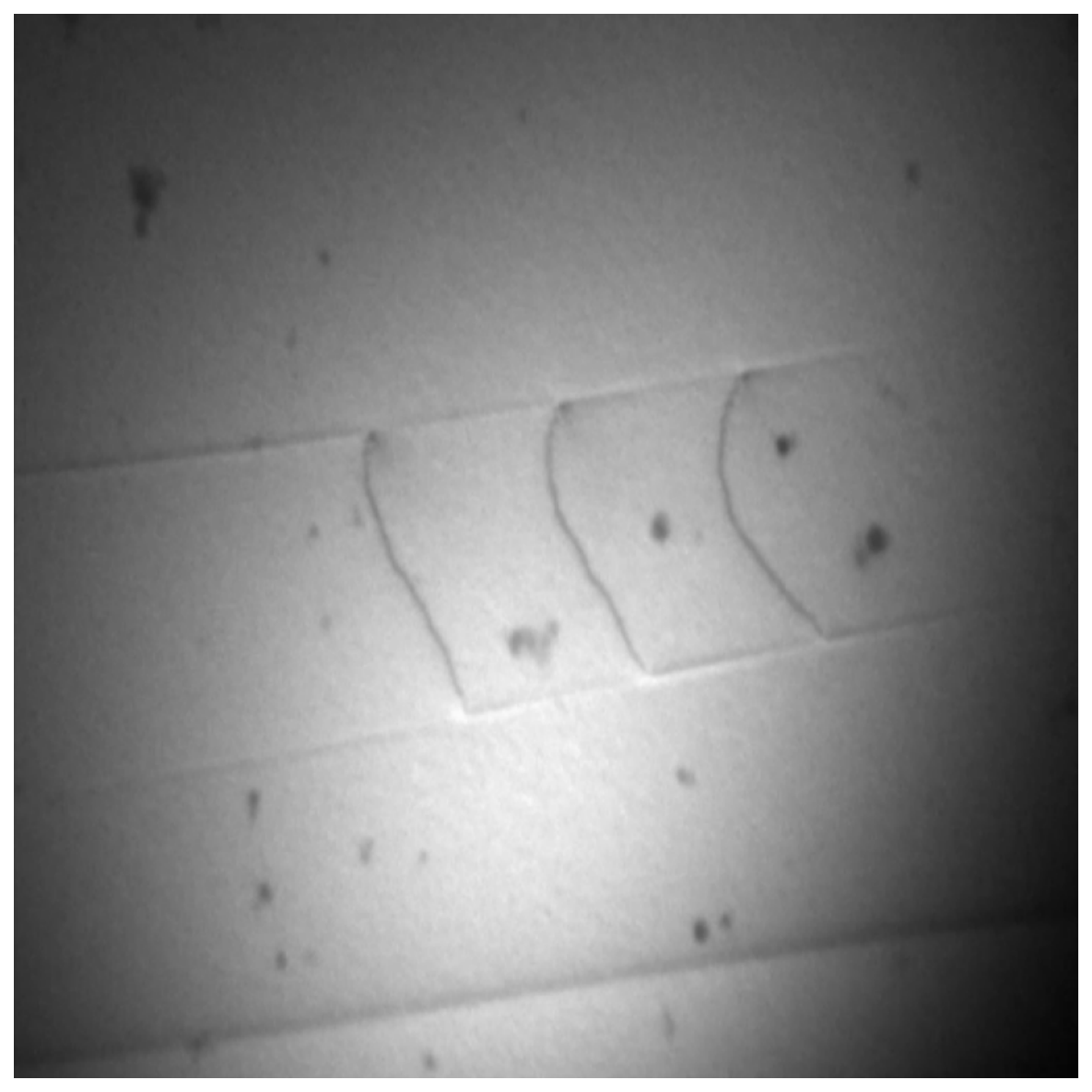}};
    \begin{scope}[x=(Bild.south east),y=(Bild.north west)]
    %\node[color=white] at (0.1,0.9) {A};
    \end{scope}
    \end{tikzpicture}}
    \hspace{-0.25cm}
    \stackinset{c}{}{t}{-.15in}{\small Segmentation}{
    \begin{tikzpicture}
    \node[anchor=south west,inner sep=0] (Bild) at (0,0)
    {\includegraphics[width=.3\columnwidth]{ 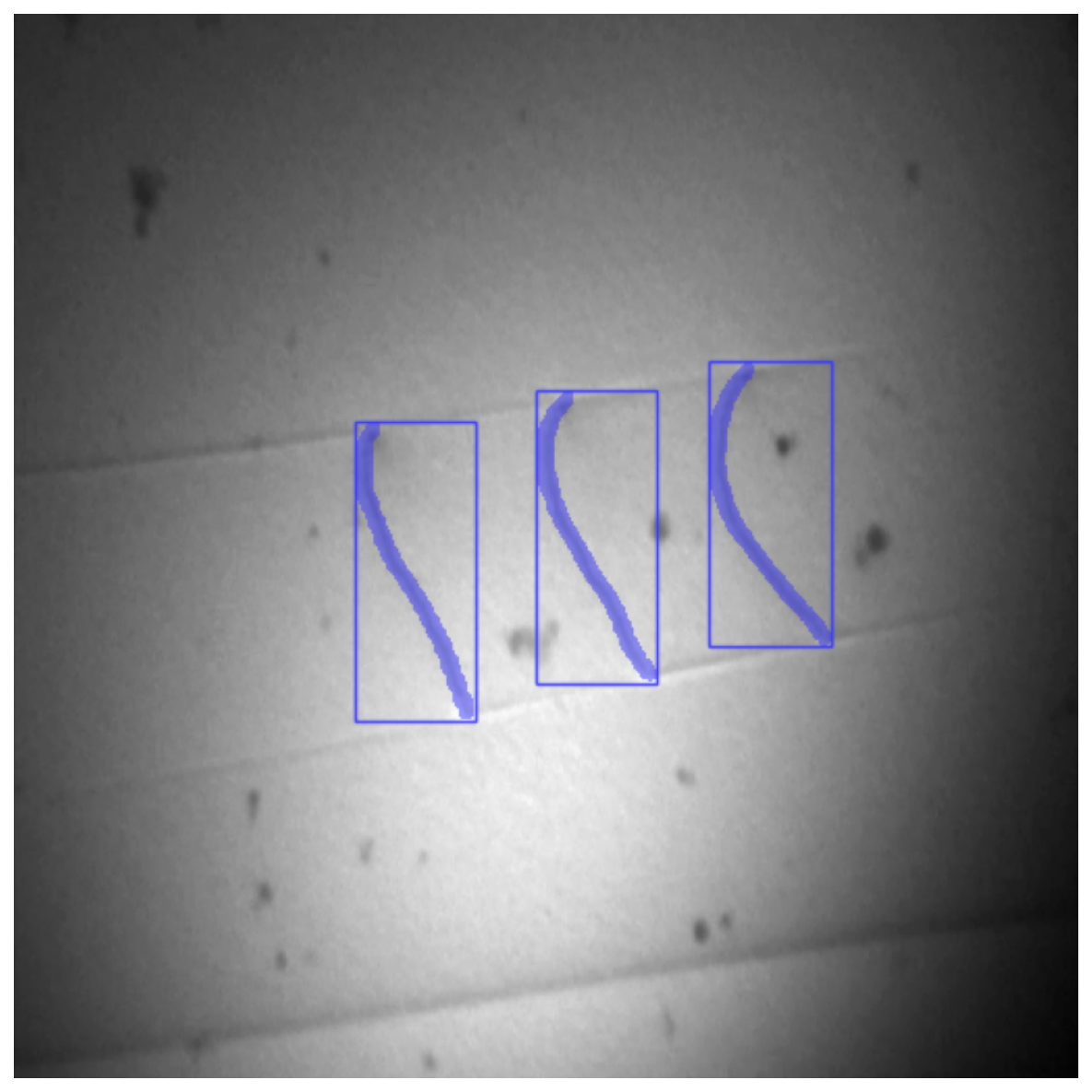}};
    \begin{scope}[x=(Bild.south east),y=(Bild.north west)]
    %\node[color=white] at (0.1,0.9) {B};
    \end{scope}
    \end{tikzpicture}}
    \hspace{-0.25cm}
    \stackinset{c}{}{t}{-.15in}{\small Skeletonized masks}{
    \begin{tikzpicture}
    \node[anchor=south west,inner sep=0] (Bild) at (0,0)
    {\includegraphics[width=.3\columnwidth]{ 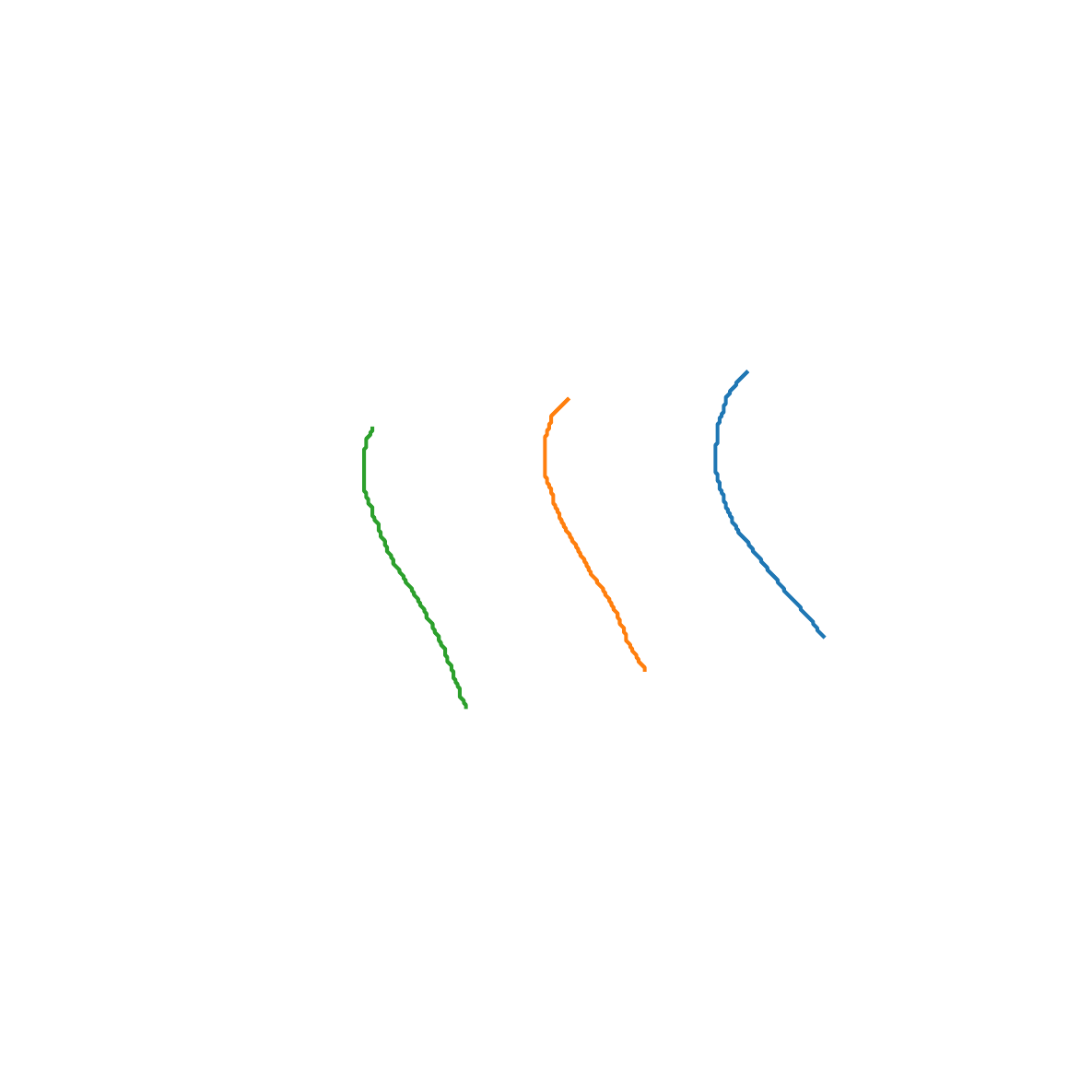}};
    \begin{scope}[x=(Bild.south east),y=(Bild.north west)]
    \node[color=black] at (0.25,0.68) {\scriptsize L$_1$=156};
    \node[color=black] at (0.6,0.28) {\scriptsize L$_2$=154};
    \node[color=black] at (0.8,0.72) {\scriptsize L$_3$=155};
    \end{scope}
    \end{tikzpicture}}
	\caption{
The proposed Dislocation segmentation pipeline that includes ML-Based segmentation, skeletonization, and the dislocation length (in pixels) extraction.
	}
	\label{fig:post processing} %fig1
\end{figure}
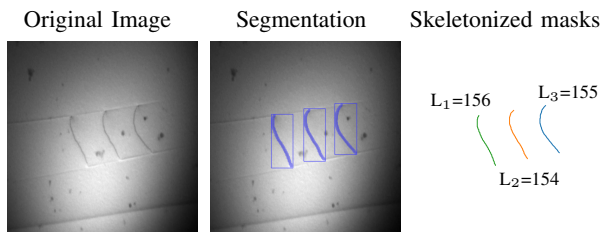

The example of skeletonized dislocations' masks, predicted using YOLOv8, along with the corresponding dislocation lengths (in pixels) is shown in Figure \ref{fig:post processing}.% by processing the instance segmentations obtained by using YOLOv8.% For visualization each Skeletonized dislocations is shown as a line using the points obtained after Skeletonization. 

\subsection{Length-aware recall}%geometry-aware recall, you name it
\label{LAR}
In order to get deeper insights into the quality of dislocation segmentation, we propose a new metric, ``Length-aware Recall'' (LAR).
%, a metric based on how accurate the models were able to segment dislocations based on their length. The pseudo code for the Length-aware recall (LAR) calculation is illustrated in Algorithm \ref{Algo}.
The input to the algorithm consists of the list of ground truth instance masks $M_{\text{gt}}$ and a list of predicted instance masks  $M_{p}$. %The algorithm calculates a metric called Length-aware recall (LAR) to measure the segmentation quality.
The proposed metric is calculated in the following way:
For each ground truth instance mask $M_{\text{gt}_i}$ in the ground truth mask list, the algorithm finds the predicted instance mask $M_{pj}$ with the highest intersection over union (IoU).
%This is done to find the predicted instance which corresponds to the true instance mask i. 
If the IoU between $M_{\text{gt}_i}$ and $M_{p_j}$ is greater than or equal to a threshold $t$ ($t=0.5$ was used in the current work), skeletonization to both masks is applied, and the lengths of their skeletons are calculated. 

We use the relative error of the dislocation lengths to calculate the value of the metric. The following three different situations are typically encountered and the metrics for them are calculated as follows:
\begin{enumerate}
    \item The predicted mask $M_{p_i}$ is solid (only one skeleton is contained in the prediction).
    In this case, the relative error $e$ between the length of the skeleton of $M_{gti}$ and that of the skeleton of $M_{p_j}$  are calculated.
    The metric value for the instance $i$ is found as a complement of the error, i.e., as $1-e$. 
    \item The predicted mask $M_{p_i}$ is fractured (several skeletons are predicted). This can be
    observed, e.g., in Figure \ref{result_im} (C) for most of the mask predictions  or for the two upper masks in Figure \ref{result_im} (0).
    Here, the length of the longest skeleton is used for the calculation.
    \item $M_{gt_i}$ did not get matched with any of the predicted masks (i.e., the dislocation was not predicted), the value LAR$_{i}$ is set to zero.
\end{enumerate}

Once the calculation has been performed for the $i$-th ground truth mask, we remove the corresponding predicted mask from the list to prevent double assignments. We then move on to the next ground truth instance and repeat the process. %For situation where the model predicts less number of instances than ground truth instances
%we set $lar$ as 0 since no predicted instance can be match with the ground truth instance. Once the $lar$ for all the
%ground truth instances has been calculated we can calculate $LAR$ for the image as the mean of all ${lar_i}$ values.

\begin{algorithm} %or another one check
 \caption{Pseudocode for the Length-aware recall (LAR) calculation}
     \SetAlgoLined
     \SetKwInOut{Input}{Input}
     \Input{
     List of ground truth instance masks: \\$M_{\text{gt}} = [M_{\text{gt}_1}, M_{\text{gt}_2}, ..., M_{\text{gt}_n}]$,\\
     List of predicted instance masks: \\$M_{p} = [M_{p_1}, M_{p_2}, ..., M_{p_m}]$,\\
     IoU threshold: $t$ (default value of 0.5)}
     \KwResult{Length-aware recall (LAR)}
     \For{ $M_{\text{gt}_i}$ \textbf{in} $M_{\text{gt}}$}{
     Find $M_{p_j}$ with the highest IoU with $M_{\text{gt}_i}$\\
     \eIf{$\text{IoU}(M_{p_i}, M_{\text{gt}_j}) \geq t $}
     {\textcolor{gray}{// Masks' skeletonization}\\
     $M^{s}_{\text{gt}_i}=\text{skeletonize}(M_{\text{gt}_i})$\\
     $M^{s}_{p_j}=\text{skeletonize}(M_{p_j})$ \\
     \textcolor{gray}{// Relative lengths' error calculation}
     $e_{i}=\left|\frac{\text{len}(M^{s}_{\text{gt}_i})-\text{len}(M^{1}_{p_j})}{\text{len}(M^{1}_{\text{gt}_i})}\right|$\\
     $\text{LAR}_i=1-e_i$
     }
     {$\text{LAR}_i=0$}
     Delete $M_{p_j}$ from $M_{p}$
    }
$\text{LAR}=\text{mean}(\text{LAR}_1, \text{LAR}_2, ...,\text{LAR}_n)$
\label{Algo}
\end{algorithm}

LAR represents the complement of the average relative error of the predicted dislocation lengths, i.e., the higher the LAR value, the better the segmentation quality of the model.

One important feature of the proposed LAR metric is that it not only penalizes false negatives, but also considers the predicted masks' quality. This means that even if a dislocation is correctly detected by the model but the predicted mask is not solid, the LAR score will reflect this and indicate that the segmentation quality is suboptimal. To accurately measure the length of the skeleton of the dislocation, the predicted mask needs to be a continuous, solid structure. If the predicted mask is fragmented or contains holes, the calculated length of the skeleton may be underestimated, resulting in a lower LAR score.

This highlights the importance of post-processing techniques to improve the quality of predicted masks, such as filling holes or smoothing edges to ensure the continuity of the structure. By incorporating these techniques, the quality of the predicted masks can be improved, in the future, which will result in a higher LAR score and a more accurate assessment of the model's performance.

\section{Results}
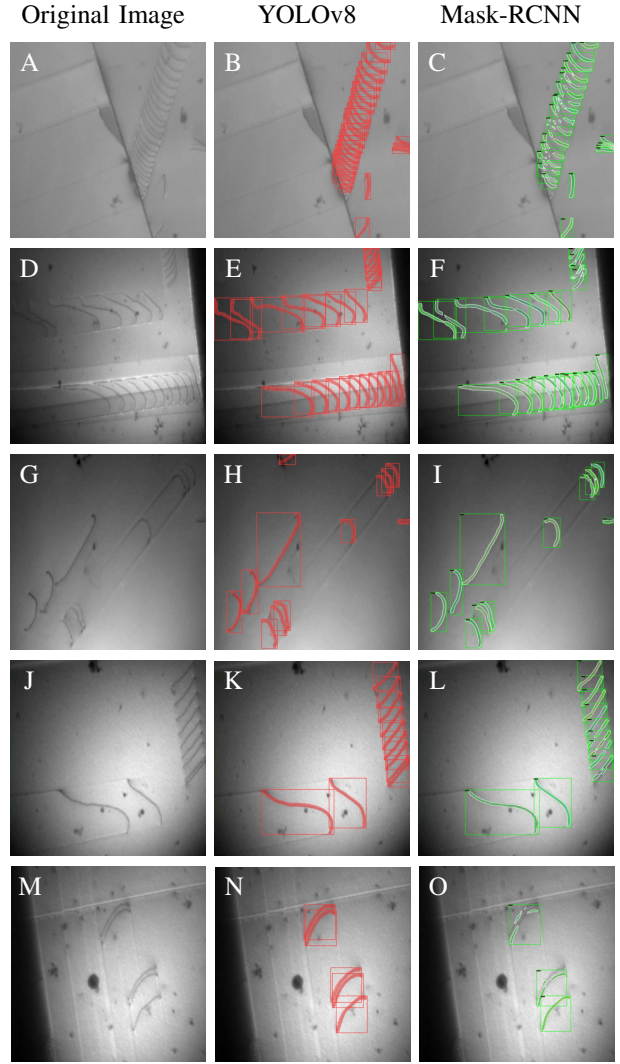
\begin{figure}[htb]
    \centering
    \hspace{-0.25cm}
    \stackinset{c}{}{t}{-.2in}{Original Image}{
    \begin{tikzpicture}
    \node[anchor=south west,inner sep=0] (Bild) at (0,0)
    {\includegraphics[width=.3\columnwidth]{ 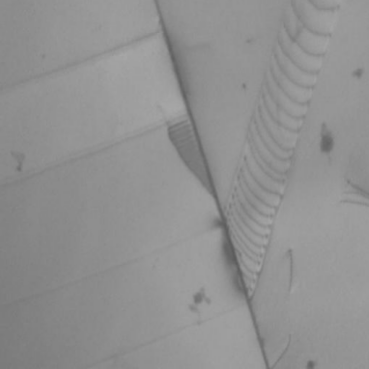}};
    \begin{scope}[x=(Bild.south east),y=(Bild.north west)]
    \node[color=white] at (0.1,0.9) {A};
    \end{scope}
    \end{tikzpicture}}
    \hspace{-0.25cm}
    \stackinset{c}{}{t}{-.2in}{YOLOv8}{
    \begin{tikzpicture}
    \node[anchor=south west,inner sep=0] (Bild) at (0,0)
    {\includegraphics[width=.3\columnwidth]{ 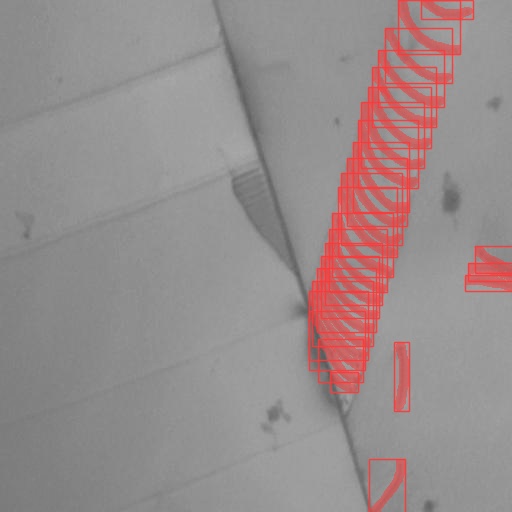}};
    \begin{scope}[x=(Bild.south east),y=(Bild.north west)]
    \node[color=white] at (0.1,0.9) {B};
    \end{scope}
    \end{tikzpicture}}
    \hspace{-0.25cm}
    \stackinset{c}{}{t}{-.2in}{Mask-RCNN}{
    \begin{tikzpicture}
    \node[anchor=south west,inner sep=0] (Bild) at (0,0)
    {\includegraphics[width=.3\columnwidth]{ 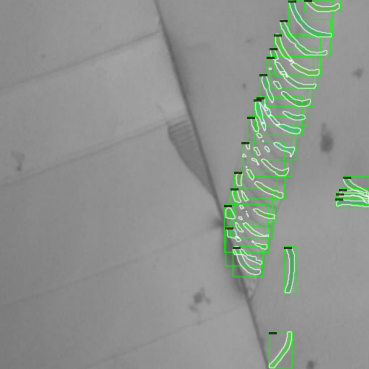}};
    \begin{scope}[x=(Bild.south east),y=(Bild.north west)]
    \node[color=white] at (0.1,0.9) {C};
    \end{scope}
    \end{tikzpicture}}
    
    \vspace{0.15cm}
    
    \begin{tikzpicture}
    \node[anchor=south west,inner sep=0] (Bild) at (0,0)
    {\includegraphics[width=.3\columnwidth]{ 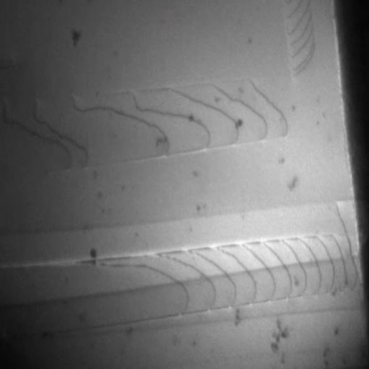}};
    \begin{scope}[x=(Bild.south east),y=(Bild.north west)]
    \node[color=white] at (0.1,0.9) {D};
    \end{scope}
    \end{tikzpicture}
    \begin{tikzpicture}
    \node[anchor=south west,inner sep=0] (Bild) at (0,0)
    {\includegraphics[width=.3\columnwidth]{ 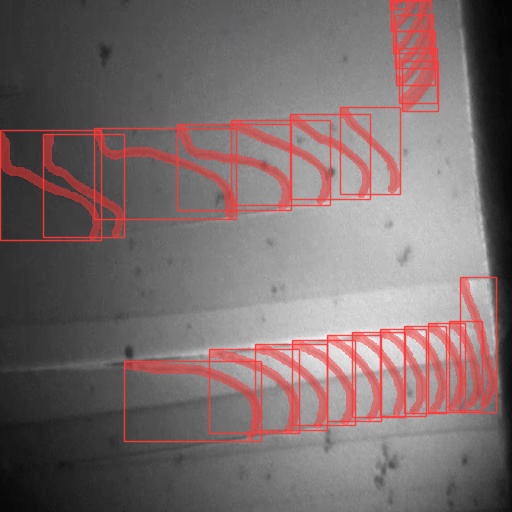}};
    \begin{scope}[x=(Bild.south east),y=(Bild.north west)]
    \node[color=white] at (0.1,0.9) {E};
    \end{scope}
    \end{tikzpicture}
    \begin{tikzpicture}
    \node[anchor=south west,inner sep=0] (Bild) at (0,0)
    {\includegraphics[width=.3\columnwidth]{ 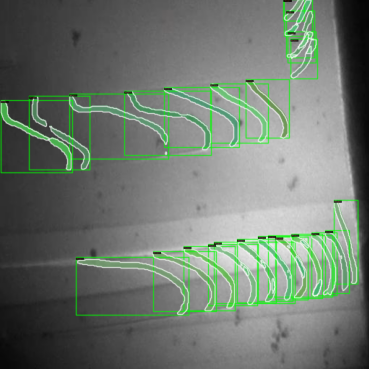}};
    \begin{scope}[x=(Bild.south east),y=(Bild.north west)]
    \node[color=white] at (0.1,0.9) {F};
    \end{scope}
    \end{tikzpicture}
    
    \vspace{0.15cm}
    \begin{tikzpicture}
    \node[anchor=south west,inner sep=0] (Bild) at (0,0)
    {\includegraphics[width=.3\columnwidth]{ 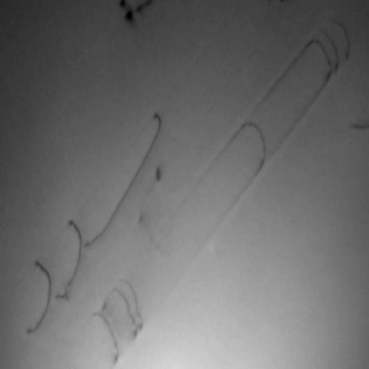}};
    \begin{scope}[x=(Bild.south east),y=(Bild.north west)]
    \node[color=white] at (0.1,0.9) {G};
    \end{scope}
    \end{tikzpicture}
    \begin{tikzpicture}
    \node[anchor=south west,inner sep=0] (Bild) at (0,0)
    {\includegraphics[width=.3\columnwidth]{ 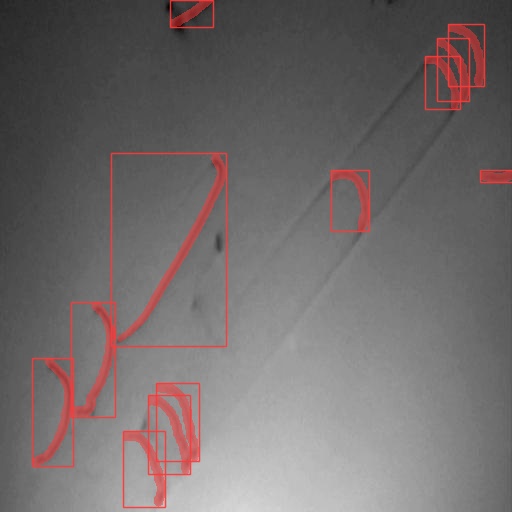}};
    \begin{scope}[x=(Bild.south east),y=(Bild.north west)]
    \node[color=white] at (0.1,0.9) {H};
    \end{scope}
    \end{tikzpicture}
    \begin{tikzpicture}
    \node[anchor=south west,inner sep=0] (Bild) at (0,0)
    {\includegraphics[width=.3\columnwidth]{ 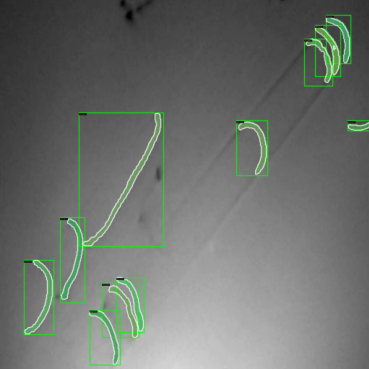}};
    \begin{scope}[x=(Bild.south east),y=(Bild.north west)]
    \node[color=white] at (0.1,0.9) {I};
    \end{scope}
    \end{tikzpicture}
    
        \vspace{0.15cm}
    \begin{tikzpicture}
    \node[anchor=south west,inner sep=0] (Bild) at (0,0)
    {\includegraphics[width=.3\columnwidth]{ 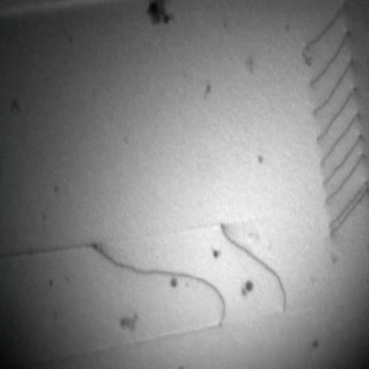}};
    \begin{scope}[x=(Bild.south east),y=(Bild.north west)]
    \node[color=white] at (0.1,0.9) {J};
    \end{scope}
    \end{tikzpicture}
    \begin{tikzpicture}
    \node[anchor=south west,inner sep=0] (Bild) at (0,0)
    {\includegraphics[width=.3\columnwidth]{ 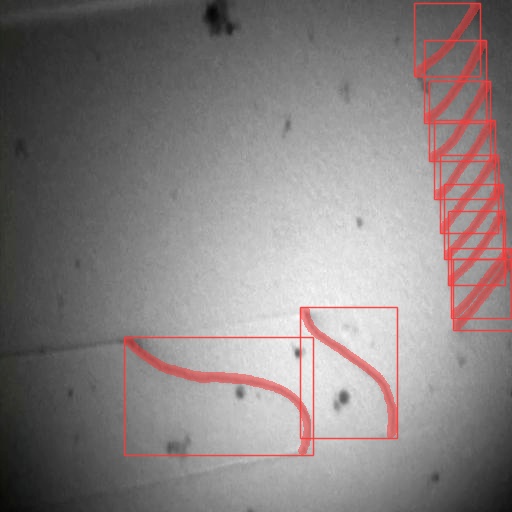}};
    \begin{scope}[x=(Bild.south east),y=(Bild.north west)]
    \node[color=white] at (0.1,0.9) {K};
    \end{scope}
    \end{tikzpicture}
    \begin{tikzpicture}
    \node[anchor=south west,inner sep=0] (Bild) at (0,0)
    {\includegraphics[width=.3\columnwidth]{ 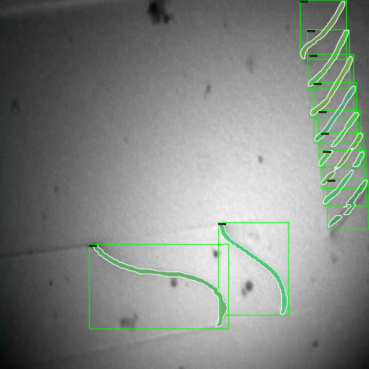}};
    \begin{scope}[x=(Bild.south east),y=(Bild.north west)]
    \node[color=white] at (0.1,0.9) {L};
    \end{scope}
    \end{tikzpicture}
    
        \vspace{0.15cm}
    \begin{tikzpicture}
    \node[anchor=south west,inner sep=0] (Bild) at (0,0)
    {\includegraphics[width=.3\columnwidth]{ 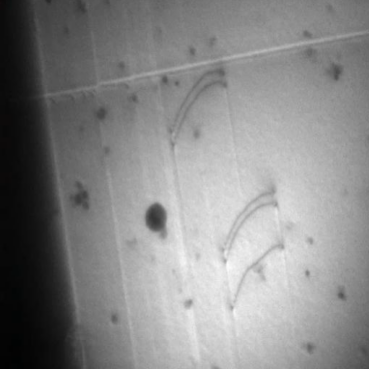}};
    \begin{scope}[x=(Bild.south east),y=(Bild.north west)]
    \node[color=white] at (0.1,0.9) {M};
    \end{scope}
    \end{tikzpicture}
    \begin{tikzpicture}
    \node[anchor=south west,inner sep=0] (Bild) at (0,0)
    {\includegraphics[width=.3\columnwidth]{ 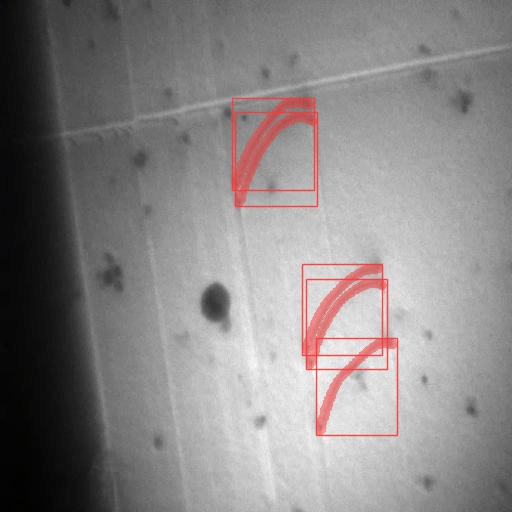}};
    \begin{scope}[x=(Bild.south east),y=(Bild.north west)]
    \node[color=white] at (0.1,0.9) {N};
    \end{scope}
    \end{tikzpicture}
    \begin{tikzpicture}
    \node[anchor=south west,inner sep=0] (Bild) at (0,0)
    {\includegraphics[width=.3\columnwidth]{ 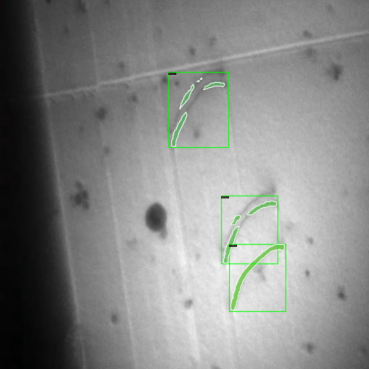}};
    \begin{scope}[x=(Bild.south east),y=(Bild.north west)]
    \node[color=white] at (0.1,0.9) {O};
    \end{scope}
    \end{tikzpicture}

    \caption{The examples of the YOLOv8 and Mask-RCNN segmentation, applied to the TEM images of the dislocations.}
    \label{result_im}
\end{figure}

Mask R-CNN and YOLOv8 networks were trained for 100 and 250 epochs, respectively. YOLOv8 provides five models of different sizes - nano (n), small (s), medium (m), large (l), and extra large (x), which were evaluated in the study. The default training parameters offered by MMDetection (Mask R-CNN) and Ultralytics (YOLOv8) were used for the networks' training process. 
Since Ultralytics offers a variety of augmentation steps with the default configuration, additional augmentation steps, i.e., 90$^\circ$ rotation, were added to the Mask R-CNN training pipeline. We evaluated the performance of the YOLOv8 instance segmentation model against Mask R-CNN on the validation set using several performance metrics, including:
\begin{itemize}
    \item the Mean Average Precision (mAP), which  is a commonly used evaluation metric in object detection and instance segmentation tasks that is calculated by computing the area under the precision-recall curve. We compute mAP for the detection and segmentation tasks. The predicted bounding box/mask is considered a true positive if it has an Intersection over Union (IoU) greater than a certain threshold. In this paper, we present the mAP$_{50}$ and the mAP5$_{50..90}$ numbers,  where mAP$_{50}$ considers multiple IoU thresholds between 0.5 and 0.9.
    \item the length-aware metric, introduced in \ref{LAR}.
\end{itemize}
The quantitative comparison of the YOLOv8 and Mask R-CNN is shown in Table \ref{results}.
\begin{table*}
  \centering
  \caption{Quantitative comparison of YOLOv8 and Mask R-CNN instance segmentation models}
  \label{results}
  \resizebox{0.65\linewidth}{!}{
  \begin{tabular}{c c c c c c c}
    \toprule 
     \multirow{2}[2]{*}{\shortstack{Framework}}&\multicolumn{2}{c}{Bounding Box}&\multicolumn{2}{c}{Mask}& \multirow{2}[2]{*}{\shortstack{LAR}}\\
     
     \cmidrule(lr){2-3} \cmidrule(lr){4-5}&mAP$_{50}$&mAP$_{50..90}$&mAP$_{50}$&mAP$_{50..90}$\\ 
    \midrule
         YOLOv8n   & 0.906& 0.720 & 0.825 & 0.374& 0.778 \\
         YOLOv8s   &  0.923& 0.758 & 0.867 & 0.413& 0.802 \\
         YOLOv8m  &  0.934& 0.768 & 0.891 & 0.442& 0.821\\
         YOLOv8l &  0.936& 0.771 & 0.901  & 0.437& 0.803 \\
         YOLOv8x &  0.937& 0.766 & 0.907  & 0.447& 0.812 \\
         Mask R-CNN& 0.805& 0.525  & 0.533 & 0.174& 0.559\\
         Mask R-CNN($w=3$)& 0.794&  0.527 & 0.572 & 0.192& 0.567\\
         Mask R-CNN($w=5$) & 0.797&  0.537 & 0.566 & 0.196& 0.555\\
         Mask R-CNN($w=10$) & 0.806&  0.531 & 0.579 & 0.203& 0.564\\
   \bottomrule
 \end{tabular}}
\end{table*}
As one can observe there, Mask R-CNN, despite being able to localize dislocations with fairly high accuracy, struggles to segment each individual dislocation.

Mask R-CNN (as well as YOLOv8) uses a multi-task loss function that combines three different losses: the classification loss, the bounding box regression loss, and the mask segmentation loss:

\begin{equation}
L=L_{\text{cls}}+L_{\text{box}}+L_{\text{mask}}
\end{equation}

Where $L_{\text{cls}}$, $L_{\text{box}}$ $L_{\text{mask}}$ are the classification, localization and segmentation losses, respectively.
In order to  increase the importance of the segmentation task and improve the segmentation accuracy of Mask-RCNN, in the training process, we increased the default weight ($w=1$) of the mask loss $ L_\text{mask}$:
\begin{equation}
L=L_{\text{cls}}+L_{\text{box}}+w\cdot L_{\text{mask}}
\end{equation}
where $w=3,5,10$ were used in the numerical experiments.
Despite the slight performance improvement caused by weighting the mask loss, all the YOLOv8 models outperformed Mask R-CNN significantly in both localization (detection) and segmentation of the dislocations, especially in the dense dislocation pile-ups, as illustrated in Figure \ref{result_im} (C, F). Moreover, as can be observed from Figure \ref{result_im}(O), the Mask-RCNN provides the "fractured" dislocation masks that would require severe post-processing, which is reflected by the relatively low LAR score (Table \ref{results}).

Besides, YOLOv8 has a user-friendly Python interface, making it more accessible for non-experts to use.

\section{Conclusions}
In this paper, we propose an automatic image-processing pipeline for identifying and localizing the dislocations in straining experiments in electron microscopy.
The proposed pipeline includes ML-based instance segmentation followed by the post-processing step of mask skeletonization.

In the context of this research, we applied and tested two state-of-the-art instance segmentation methods. The comparative analysis revealed that the YOLOv8 framework shows a better performance for dislocation segmentation.

Additionally, we proposed a novel geometry-aware segmentation metric that provides deeper insights into the quality of the segmentation results. This metric was used to evaluate the accuracy of segmentation methods with regard to the dislocation length and geometry.

The proposed pipeline saves time and resources for material scientists who analyze in-situ TEM straining experiments, and it eliminates the need for manual time-consuming analysis, leading to more efficient and productive research.

The source code for the proposed Length-aware Recall calculation and the pre-trained YOLOv8 models are available on 
%GitHub \url{https://github.com/kruzaeva/dislocation-segmentation}.
Zenodo \cite{zenodo}. A repository containing updates of the code and data can be found at \cite{gitlab}.

\section{Acknowledgment}
This work was funded by the European Research Council through the ERC Grant Agreement no. 759419 MuDiLingo (``A Multiscale Dislocation Language for Data-Driven Materials''). 
The authors would like to thank Prof. Antonin Dlouhy (IPM Czech Academy of Science, Brno) for providing the Cantor alloy samples that were deformed inside the TEM.
\addtolength{\textheight}{-12cm}

\bibliographystyle{IEEEbib}
\bibliography{bibliography}

\end{document}